\documentclass[lettersize,journal]{IEEEtran}
\usepackage{amsmath,amsfonts}
\usepackage{algorithmic}
\usepackage{algorithm}
\usepackage{array}
\usepackage[caption=false,font=normalsize,labelfont=sf,textfont=sf]{subfig}
\usepackage{textcomp}
\usepackage{booktabs}
\usepackage{mathtools}  
\usepackage{array}      
\usepackage{siunitx}
\usepackage{multirow}
\usepackage{stfloats}
\usepackage{url}
\usepackage{verbatim}
\usepackage{amssymb}
\usepackage{bbm} 
\usepackage{graphicx}
\usepackage{cite}
\usepackage{xcolor}
\usepackage{tcolorbox}
\usepackage{colortbl}

\begin{document}

\title{REE-TTT: Highly Adaptive Radar Echo \\Extrapolation Based on Test-Time Training}

\author{Xin Di,
        Xinglin Piao, \IEEEmembership{Member, IEEE}, 
        Fei Wang, 
        Guodong Jing, 
        Yong Zhang, \IEEEmembership{Member, IEEE}
\thanks{This work was supported in part by the Key Laboratory of Smart Earth (Grant No. KF2023ZD03-05); the CMA Innovative and Development Program (Grant No. CXFZ.20231035); the National Key R\&D Program of China (Grant No. 2021ZD0111902); the National Natural Science Foundation of China (Grant Nos. 62472014 and U21B2038); and the Scientific and Technological Project of China Meteorological Administration (Grant No. CMAJBGS202505). (\emph{The corresponding author is Yong Zhang.)}}
\thanks{Xin Di, Xinglin Piao and Yong Zhang are with Beijing Key Laboratory of Multimedia and Intelligent Software Tech nology, Faculty of Information Technology, Beijing Institute of Artificial
Intelligencea(e-mail: CLMWJY3@163.com;  piaoxl@bjut.edu.cn; zhangyong2010@bjut.edu.cn)}
\thanks{Fei Wang and Guodong Jing are with China Meteorological Administration Weather Modification Center, Beijing 100081, China (e-mail:  feiwang@cma.cn; Jinggd@cma.cn).}}

\markboth{IEEE TRANSACTIONS ON GEOSCIENCE AND REMOTE SENSING,~Vol.~XX, No.~X, August~2025}%
{Di \MakeLowercase{\textit{et al.}}: REE-TTT: Highly Adaptive Radar Echo Extrapolation Based on Test-Time Training}

\maketitle

\begin{abstract}
Precipitation nowcasting is critically important for meteorological forecasting. Deep learning-based Radar Echo Extrapolation (REE) has become a predominant nowcasting approach, yet it suffers from poor generalization due to its reliance on high-quality local training data and static model parameters, limiting its applicability across diverse regions and extreme events. To overcome this, we propose REE-TTT, a novel model that incorporates an adaptive Test-Time Training (TTT) mechanism. The core of our model lies in the newly designed Spatio-temporal Test-Time Training (ST-TTT) block, which replaces the standard linear projections in TTT layers with task-specific attention mechanisms, enabling robust adaptation to non-stationary meteorological distributions and thereby significantly enhancing the feature representation of precipitation. Experiments under cross-regional extreme precipitation scenarios demonstrate that REE-TTT substantially outperforms state-of-the-art baseline models in prediction accuracy and generalization, exhibiting remarkable adaptability to data distribution shifts.
\end{abstract}


\section{Introduction}
\IEEEPARstart{W}{ith} the rapid advancement of deep learning, data-driven weather prediction methods have gained significant traction, particularly in Radar Echo Extrapolation (REE) for high-resolution precipitation nowcasting. These approaches leverage historical radar sequences to model spatio-temporal patterns of precipitation systems, outperforming traditional methods in short-term forecasting. This technological progress coincides with unprecedented climate challenges: 2024 was recorded as the hottest year in history by the World Meteorological Organization. Regions with traditionally stable climatic regimes are now experiencing unprecedented weather anomalies, as exemplified by the Mediterranean storm of 2023 that induced catastrophic flooding in Libya, a region historically characterized by arid conditions. This underscores the urgent need to establish accurate and universal weather forecasting systems, particularly to develop adaptive prediction frameworks capable of maintaining accuracy in evolving climate scenarios.

Traditional meteorological prediction relies on Numerical Weather Prediction (NWP). This approach constructs differential equation systems based on atmospheric physics principles to model atmospheric motion, then numerically solves these equations to forecast future states~\cite{ref1}. Although effective for mid-to-long-term weather trend forecasting and large-scale system evolution analysis~\cite{ref2}, NWP's heavy dependence on initial conditions and computational complexity result in considerable time lags, limiting its capability to resolve small-scale weather systems. Consequently, NWP struggles to meet the critical requirements of short-term precipitation nowcasting: minute-level update frequency and kilometer-level spatial resolution~\cite{ref3}.

In contrast, REE methods analyze radar data to infer future precipitation fields. This approach enables high-resolution nowcasting within shorter time frames while capturing the evolutionary characteristics of meso- and micro-scale weather systems~\cite{ref4}. Classical REE techniques include centroid tracking~\cite{ref5, ref6, ref7}, cross-correlation~\cite{ref8, ref9, ref10}, and optical flow methods~\cite{ref11, ref12, ref13}. The centroid tracking method calculates the centroid positions of radar echo clusters to estimate their trajectories. Cross-correlation methods determine optimal displacement vectors by measuring spatial similarity between consecutive radar images. Optical flow approaches estimate pixel-level velocity fields through brightness variation analysis. However, these methods presuppose continuous and smooth motion patterns of radar echoes, whereas precipitation events often exhibit intense convection bursts, morphological mutations, and structural dissipation that challenge traditional REE paradigms.

\begin{figure}[!t]
\centering
\includegraphics[width=0.9\columnwidth]{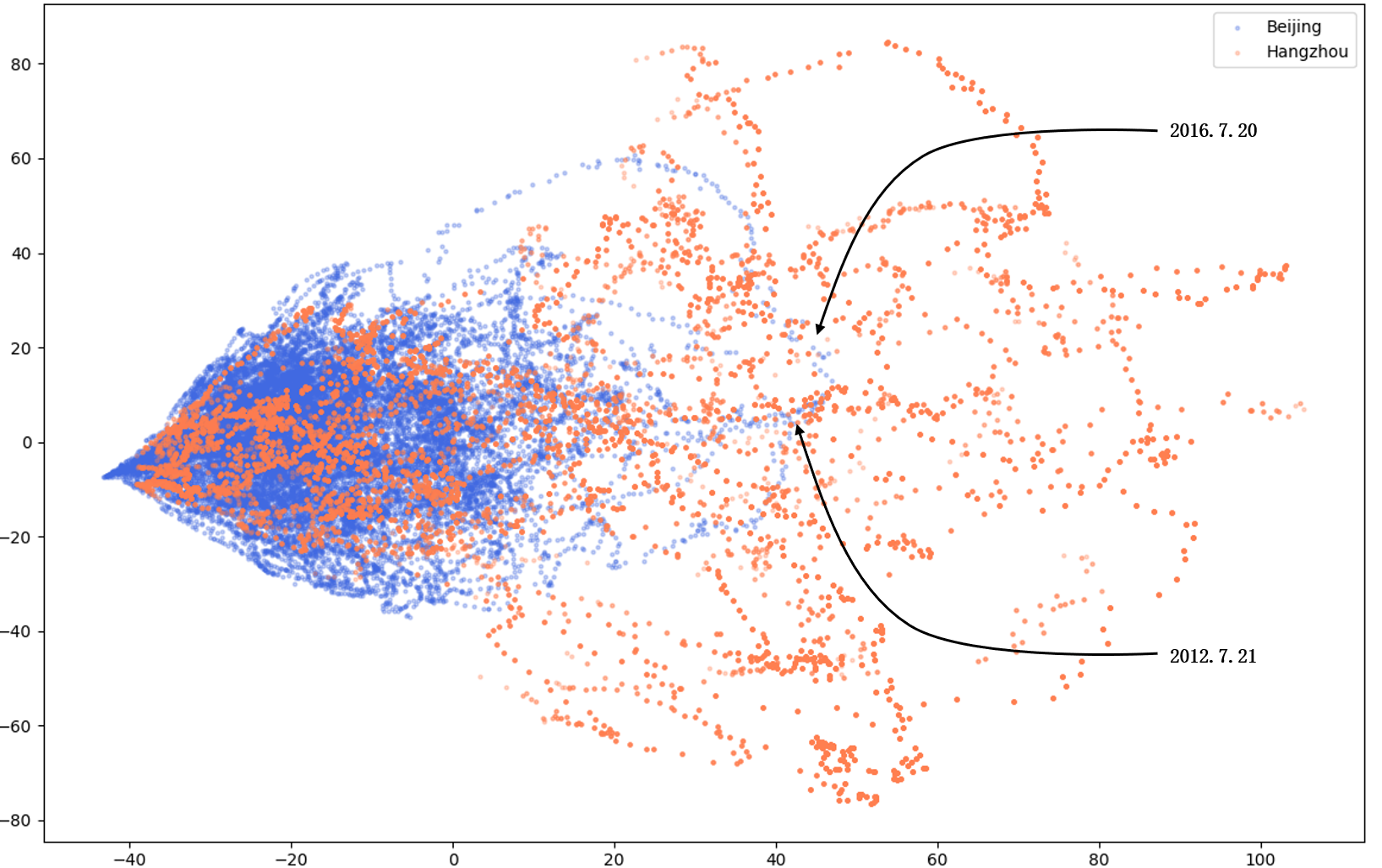} 
\caption{Cluster analysis results of radar composite reflectivity samples from the Beijing and Hangzhou datasets reveals distinct distributions, where each point corresponds to a radar composite reflectivity image. Most samples in both datasets reside near cluster centers, representing low-intensity precipitation patterns. However, the Hangzhou dataset contains a greater number of outlier samples corresponding to intense precipitation events. In contrast, while the Beijing dataset is dominated by low-intensity  precipitation, it also includes several outlier processes that deviate from these clusters, particularly samples captured during two historic torrential rain events (marked in the figure).}
\label{fig_1}
\end{figure}

Deep learning technologies offer new solutions to overcome these limitations. Leveraging the non-linear modeling capabilities of deep neural networks, end-to-end learning mechanisms can automatically extract spatio-temporal evolution features from historical radar echo sequences for predictive modeling. Recent years have witnessed remarkable progress in deep learning-based REE research, including models based on recurrent neural networks (RNNs)~\cite{ref21, ref22, ref24, ref25, ref27}, attention-enhanced architectures~\cite{ref31, ref32, ref33, ref34} and generative approaches~\cite{ref36, ref37}.

Despite their demonstrated success, these models employ static training paradigms under fixed meteorological scenarios, achieving competent performance within specific training data distributions. However, in practical applications, constrained by historical precipitation records, acquiring high-quality regional observational data remains challenging, hindering these models' ability to adapt to divergent meteorological conditions. As shown in Fig.~\ref{fig_1}, the cluster analysis reveals significant distributional discrepancies between various regions and different precipitation processes. This poses two critical limitations on existing models: 

\textbf{Inability for Cross-region Deployment:}
Traditional models typically require strict distribution alignment between training and testing data, thus necessitating localized retraining when deployed in new geographical domains. However, training, optimizing, and maintaining separate models for each radar station incurs high costs. Furthermore, in regions where historical precipitation data are insufficient, obtaining high-quality training data presents a significant challenge, making it difficult for these areas to effectively utilize existing meteorological models for high-resolution native precipitation nowcasting.

\textbf{Inability for Adaptation to Unforeseen Precipitation Events:}
In narrow historical training datasets, samples capturing extreme weather events are typically insufficient, resulting in inadequate model learning when encountering such scenarios. Consequently, static models exhibit poor adaptability during extreme weather events. When faced with weather patterns beyond the training data distribution, such as abrupt torrential rain or hailstorms, model accuracy often degrades significantly.

To address these challenges, we propose a Radar Echo Extrapolation model incorporating Test-Time Training (REE-TTT). Departing from static models, REE-TTT dynamically adjusts its feature representations based on incoming real-time radar sequences through self-supervised learning. This enables it to effectively capture the non-stationary spatio-temporal patterns characteristic of weather events, enhancing its adaptability when encountering unforeseen scenarios. Crucially, this paradigm shift significantly strengthens the model's cross-regional generalization capabilities. We summarize the key contributions as follows:
\begin{itemize}
\item We propose REE-TTT, the first radar echo extrapolation model integrating test-time training (TTT) layers, enabling dynamic parameter adjustment during inference to address distribution shifts across meteorological scenarios.

\item We design an attention-enhanced spatio-temporal test-time training (ST-TTT) block that first integrates task-specific attention mechanisms to reconstruct feature views, replacing the standard linear projections in traditional TTT layers for efficient spatio-temporal dependency capturing.

\item Benchmark evaluations on Beijing composite reflectivity datasets and zero-shot generalization experiments on Hangzhou heavy precipitation scenarios demonstrate REE-TTT's superior robustness and adaptability.
\end{itemize}

\section{Related Work}
This section reviews recent advancements in two critical technologies supporting our work: spatio-temporal prediction neural networks and test-time training.

\subsection{Spatio-temporal Prediction}
Deep learning-based radar echo extrapolation tasks are typically modeled through spatio-temporal prediction approaches, where the predominant methodology integrates convolutional and recurrent architectures to leverage their respective strengths in spatial and temporal modeling. ConvLSTM~\cite{ref21} replaces fully-connected operations in traditional Long Short-Term Memory (LSTM) networks with convolutional layers, effectively capturing spatio-temporal dependencies and demonstrating superior performance in radar echo extrapolation. Building upon this foundation, PredRNN~\cite{ref22} and its enhanced variants~\cite{ref23} introduce vertical hidden state transitions within recurrent structures, significantly improving long-term sequence prediction capabilities. The Memory In Memory (MIM) network~\cite{ref24} innovates through differential memory storage mechanisms across timesteps, optimizing feature fusion in RNNs. Other works adopting similar stacked spatio-temporal LSTM frameworks include~\cite{ref25, ref26}, which replace convolutions with attention modules to achieve global receptive fields. Alternative approaches employ convolutional networks for spatial feature extraction followed by recurrent networks for temporal evolution modeling, as seen in~\cite{ref27, ref28}.

While RNNs remain central to temporal dynamics modeling in the aforementioned architectures, their high computational costs motivate alternative designs. SIMVP~\cite{ref29} pioneers a paradigm shift by implementing spatio-temporal feature fusion through lightweight convolutional translators, substantially reducing training overhead. This inspires subsequent innovations: TAU~\cite{ref30} captures spatio-temporal dynamics via intra-frame self-attention and inter-frame cross-attention fusion, while Earthfarseer~\cite{ref31} establishes a parallel Transformer-convolutional architecture enhanced by Fourier transforms, enabling synergistic modeling of global dependencies and local interactions. Other no-recurrent approaches employ stacked Transformer blocks~\cite{ref32, ref33, ref34} and 3D wavelet transforms~\cite{ref35}, achieving efficient modeling through spatio-temporal feature reconstruction while maintaining low computational complexity. Generative approaches like diffusion models~\cite{ref36, ref37} produce detailed predictions through iterative refinement, though this paradigm fall beyond our research scope.

In summary, current deep spatio-temporal prediction methodologies bifurcate into two technical lineages: RNN-based models, which prioritizes temporal dependency capture at the cost of training complexity; Non-Recurrent models, that enable multi-scale feature interaction through parallelizable modules like convolutional translators and spatio-temporal attention, albeit potentially sacrificing fine-grained temporal dynamics.

\subsection{Test-Time Training}
The concept of training on test data originates from local learning in Support Vector Machines (SVMs)~\cite{ref15} and transductive learning~\cite{ref16}. Both paradigms leverage unlabeled data: local learning trains task-specific models within neighborhoods of test inputs to enhance predictive capability through local information, while transductive learning directly constrains decision boundaries using unlabeled test samples to optimize performance on specific test sets.

In deep learning, conventional training paradigms optimize model parameters solely on training data and apply the static model to all test samples during inference. However, such fixed models suffer significant performance degradation under distribution shifts between training and test data. To address this challenge, TTT extends model optimization into the inference phase by creating sample-specific generalization tasks for each test instance. This allows dynamic parameter adjustment based on test data characteristics to improve predictions.

The core challenge of TTT lies in designing appropriate generalization tasks. These tasks must satisfy two criteria: self-supervision due to the absence of test labels and generalizability to extract key features from arbitrarily distributed test samples. Common self-supervised tasks such as rotation prediction~\cite{ref17} and image patch masking reconstruction~\cite{ref18} have been validated for TTT frameworks~\cite{ref19, ref20}. The TTT layer~\cite{ref14} implements this strategy through learned reconstruction tasks - unlike methods relying on predefined human priors, its reconstruction objectives are autonomously acquired during model training. This eliminates manual task engineering constraints and enables more universal task representations.

Previous studies have demonstrated the feasibility of TTT strategies in addressing complex tasks. Building upon this foundation, our model addresses the prevalent underutilization of test sequence information in existing radar echo extrapolation models by employing an attention-enhanced network to learn self-supervised reconstruction tasks within TTT layers.

\section{METHODOLOGY}
In this section, we elaborate on the proposed REE-TTT,  Section~\ref{sec:problem} defines the radar echo extrapolation task, followed by Section~\ref{sec:overview} presenting the overall architecture and key components of the REE-TTT model. Section~\ref{sec:ST-TTT} focuses on the core ST-TTT block, while Section~\ref{sec:loss} concludes with the formulation of the model's loss function.
\subsection{Problem Formulation}\label{sec:problem}
We first formally define the radar echo extrapolation task. Given a historical radar echo image sequence of length $T$ at the current time $t$, denoted as $\mathcal{X}^{(t,T)}=\{x_{i}\}_{i=t-T+1}^{t}\in\mathbb{R}^{T\times C\times H\times W}$, where the channel number $C=1$ (representing the radar reflectivity factor in dBZ), and $H\times W$ denotes the spatial dimensions. The model aims to learn parameters $\theta$ of mapping function $F_{\theta}$ that predicts a future radar echo sequence of length $T^{\prime}$, denoted as $\mathcal{Y}^{(t+1,T^{\prime})}=\{y_{i}\}_{i=t+1}^{t+T^{\prime}}\in\mathbb{R}^{T^{\prime}\times C\times H\times W}$. The optimization objective is formulated as:
\begin{equation}
\label{equation1}
\theta^{*}=\arg\min_{\theta } \mathcal{L}\big(F_{\theta}(\mathcal{X}^{(t,T)}),\mathcal{Y}^{(t+1,T^{\prime})}\big),
\end{equation}
where $\mathcal{L}(\cdot)$ denotes the composite loss function quantifying prediction errors, and $\theta^{*}$ represents the optimal parameter set for the radar echo extrapolation model.

\begin{figure*}[!t]
\centering
\includegraphics[width=1.8\columnwidth]{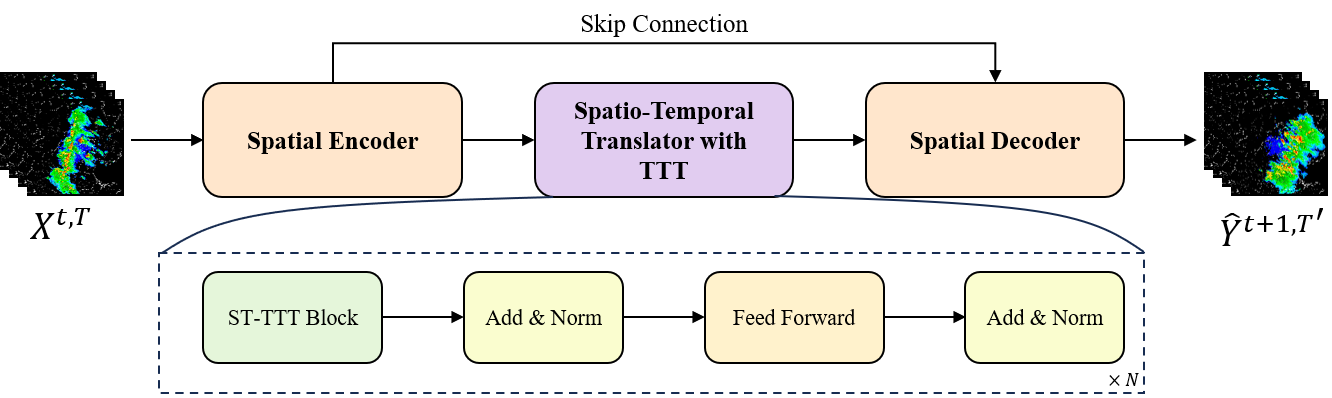}
\caption{Overview of REE-TTT Model.}
\label{fig_2}
\end{figure*}

\subsection{Overview}\label{sec:overview}
We construct the REE-TTT model to learn the mapping $F_{\theta}$, with its overall architecture illustrated in Fig. \ref{fig_2}. Following the streamlined framework proposed by SIMVP~\cite{ref29}, it comprises three core modules: a spatial encoder, a spatio-temporal translator, and a spatial decoder. This framework provides a proven and efficient backbone for spatio-temporal feature processing, whose clear separation between spatial encoding/decoding and temporal translation in the translator component offers an ideal architectural basis for integrating our novel TTT mechanism. The spatial encoder extracts frame-wise spatial-static features from radar echoes; the spatio-temporal translator integrates temporal-dynamic features into the embedded spatial representations; and the spatial decoder reconstructs high-dimensional features into predicted image sequences. Additionally, we design skip connections that directly incorporate bottom-layer features into predictions, as detailed below:

\textbf{Spatial Encoder-Decoder}: A symmetric encoder-decoder architecture constructs a multi-scale spatio-temporal feature pyramid. The encoder employs stacked 2D $3\times3$ convolutions with an independent temporal-step processing strategy to preserve spatio-temporal feature independence. It maps raw images into feature space through progressive spatial downsampling, where low-level features retain fine-grained spatial structures of radar echoes, while high-level features encode global semantic information. The decoder consists of transposed convolutions symmetric to the encoder, forming an inverse feature mapping process. To mitigate the detail loss and averaging effects inherent in standard upsampling operations, we augment the decoder with a super-resolution branch inserted after the upsampling layers. This branch, composed of multiple Residual in Residual Dense Blocks~\cite{ref42}, takes the upsampled features as input and learns to generate high-frequency residuals that refine the precipitation cores, thereby enhancing the sharpness and accuracy of predictions. The final feature fusion module adopts a channel-wise weighted fusion strategy to integrate skip-connected low-level detail features, decoder-generated high-level semantic features, and the super-resolution refined features for joint reconstruction of predicted sequences.

\textbf{Spatio-temporal Translator with TTT}: Positioned between the encoder's feature compression layer and the decoder's reconstruction layer, this module first reorganizes the encoder's multi-channel discrete spatio-temporal feature tensor by merging channel and temporal dimensions. This projects discrete features into a continuous spatio-temporal flow, generating dynamically coherent tensor representations. The feature flow undergoes deep transformation through cascaded residual TTT blocks, each containing layer normalization, an ST-TTT block, a feedforward module, and residual connections. As shown in the lower half of Fig. \ref{fig_2}, this design enables efficient spatio-temporal feature extraction. Crucially, the incorporated TTT strategy allows adaptive model adjustments based on varying precipitation evolution patterns, with ST-TTT implementation details provided in Section III-C.

\textbf{Skip Connections}: The evolutionary characteristics of radar echoes adhere to inherent physical motion laws. These inherent features do not require task-specific self-supervised learning. Thus, the skip connection path employs a parallel attention block combining motion attention~\cite{ref38} and temporal attention~\cite{ref30}, replacing the TTT block in Spatio-temporal Translator. This branch directly extracts general motion patterns and evolutionary characteristics from raw radar echoes.By integrating these essential characteristics, skip connections ensure the backbone network focuses on domain‑specific spatio‑temporal patterns induced by distribution shifts while maintaining the fidelity of local structures, thereby achieving targeted adaptation and improved prediction quality.

\begin{figure}[!t]
\centering
\includegraphics[width=0.9\columnwidth]{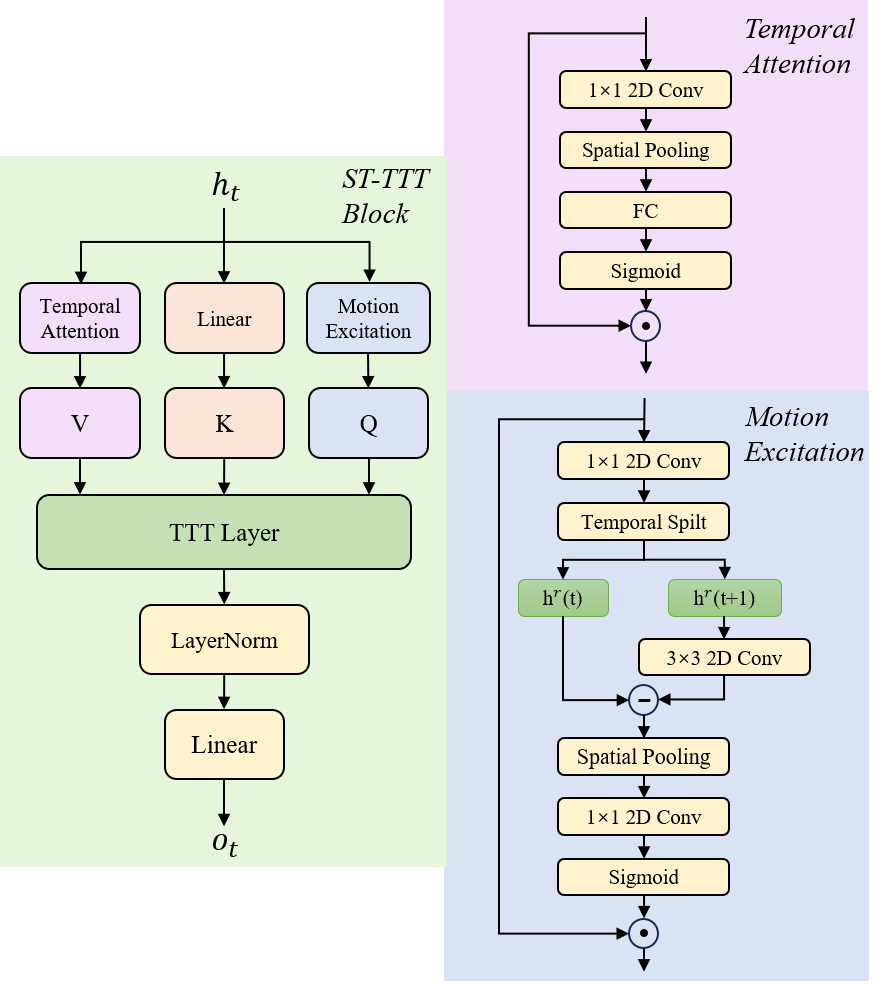}
\caption{Proposed ST-TTT block.}
\label{fig_3}
\end{figure}
\subsection{Spatio-Temporal TTT block}\label{sec:ST-TTT}
Based on the TTT layers and leveraging inherent characteristics of spatio-temporal prediction tasks, we design the ST-TTT block as the core component of the spatio-temporal translator. Let $h\in\mathbb{R}^{B\times T\times C^{\prime}\times H^{\prime}\times W^{\prime}}$ denote the encoded spatial features sequence, where $C^{\prime}, H^{\prime}, W^{\prime}$ are the feature dimensions obtained by downsampling through the encoder convolution. The ST-TTT module extracts evolutionary features from $h$ and processes them into predictive features. As illustrated in Fig. \ref{fig_3}, the architecture adopts a Transformer backbone embedded with TTT layers, structured as a dual-loop training framework:

\textbf{Outer Loop}: Performs global parameter updates through supervised learning as defined in Section III-A. The parameter set $\theta$ in Eq.~\eqref{equation1} includes critical global parameters $\theta _{q}, \theta _{k}, \theta _{v}$ of the TTT layer, which are updated via the outer loop. These parameters project inputs to distinct views to formulate self-supervised reconstruction tasks for the inner loop:

\begin{equation}
\label{equation2}
\theta_{q},\theta_{k},\theta_{v}\leftarrow\theta_{q}-\eta_{\mathrm{out}}\nabla\mathcal{L},\quad\theta_{k}-\eta_{\mathrm{out}}\nabla \mathcal{L},\quad\theta_{v}-\eta_{\mathrm{out}}\nabla\mathcal{L}.
\end{equation}

\textbf{Inner Loop}: The ST-TTT block enables dynamic parameter adaptation during inference by learning a self-supervised model $f_{\mathrm{in}}$ under a reconstruction loss $l_{\mathrm{in}}$. This core mechanism allows the model to tailor its parameters to the specific characteristics of each test sequence.

The adaptation process proceeds as follows: at each optimization step $n$, the self-supervised task is defined as reconstructing the label view from the training view:
\begin{equation}
\label{equation3}
l_{\mathrm{in}}(h;W_n) = \left| f_{\mathrm{in}}(h\theta_{k}, W_n) - h\theta_{v} \right|^{2},
\end{equation}
where $\theta_{k}$ and $\theta_{v}$ represent the fixed projection matrices for training and label views respectively, learned during the outer-loop training. 

The parameters $W$ of the inner-loop model $f_{\mathrm{in}}$ are iteratively updated via gradient descent. Starting from initialization $W_0$, the update at step $n$ is:
\begin{equation}
\label{equation4}
W_{n}=W_{n-1}-\eta_{\mathrm{in}}\nabla{l}_{\mathrm{in}}(W_{n-1},h),
\end{equation}
where $\eta_{\mathrm{in}}$ is the inner-loop learning rate. This optimization employs a spatial-major strategy, enabling efficient gradient-based adaptation to local precipitation dynamics. Crucially, this inner-loop optimization persists during testing, allowing real-time adaptation to distribution shifts. 

Finally, the output feature is computed by projecting the input onto the test view $h\theta_{q}$ and processing it with updated inner-loop model:
\begin{equation}
\label{equation5}
o=f_{\mathrm{in}}\bigl(h\theta_{q},W_{N}\bigr).
\end{equation}

Standard TTT layers implement $\theta_{q}, \theta_{k}, \theta_{v}$ through linear projections. However, for radar echo extrapolation tasks, such linear projection schemes lack task-specific focus, fail to capture localized spatio-temporal patterns, and cannot explicitly model dynamic temporal correlations. To address these limitations, we propose an attention-enhanced feature projection scheme formulated as follows:

\begin{align}
\text{Label view:} & \quad h^V = \mathrm{TA}(h; \theta_v), \\
\text{Training view:} & \quad h^K = \mathrm{Linear}(h;\theta_k), \\
\text{Test view:} & \quad h^Q = \mathrm{ME}(h; \theta_q).
\end{align}

This design replaces the standard linear projections with dedicated attention mechanisms. Specifically, we employ well-established, general-purpose spatio-temporal attention modules (TA and ME) for their computational efficiency and parallelizability, which are essential for supporting the frequent inner-loop optimizations required during test-time inference, thereby meeting real-time forecasting constraints. Moreover, their universal design enables the model to learn adaptable representations from the data, which promotes robust generalization across diverse precipitation systems and geographical regions—a core advantage of the TTT adaptation strategy.

$\mathrm{Linear}(h;\theta_k)$ applies linear projection with weights $\theta_k$, producing base features containing both discriminative spatio-temporal patterns. This serves as input for the reconstruction task.

$\mathrm{TA}(h; \theta_v)$ denotes the Temporal Attention Module parameterized by $\theta_v$. The module employs a squeeze-and-excitation mechanism that first applies global average pooling over spatial dimensions to aggregate temporal statistics, generating a time-channel-wise descriptor. This descriptor is then processed through a feed-forward network with a reduction layer and nonlinear activation, which learns to emphasize time channels that exhibit significant temporal variations. By adaptively recalibrating time-channel importance through learned attention weights, the module enhances features that represent critical temporal transitions, thus effectively capturing the evolution patterns of precipitation systems.

$\mathrm{ME}(h; \theta_q)$ represents the Motion Enhanced Attention module parameterized by $\theta_q$. It enhances the dynamic evolution patterns of $h$, establishing the reconstruction target to strengthen dynamic perception of the sequence and providing a query view infused with real-time motion cues during testing. Under the optimized model $f_{\mathrm{in}}$, the motion-aware query view enables the output feature $o$ to integrate both learned temporal evolution patterns and instantaneous motion clues. This functional decoupling of feature projections, achieved through distinct attention mechanisms, significantly enhances the representational capacity of the spatio-temporal translator.

This functional decoupling of feature projections into distinct attention mechanisms forces $f_{\mathrm{in}}$ to learn associations between basic echo patterns and their spatio-temporal evolution through the self-supervised task. Consequently, the motion-aware query view enables the output feature $o$ to integrate both learned temporal patterns and instantaneous motion clues. This design significantly enhances the representational capacity of the spatio-temporal translator, allowing the model to continuously refine its understanding of precipitation evolution during inference, making it particularly effective for handling unforeseen weather scenarios and cross-regional distribution shifts.

\subsection{Loss Functions}\label{sec:loss}
Spatio-temporal prediction tasks typically employ pixel-wise error metrics for loss $\mathcal L$. However, radar echo images contain extensive echo-free or weak-echo regions, and direct pixel-wise loss calculation tends to over-smooth strong echo areas, leading to blurred predictions. Given the critical need for extreme precipitation warnings, we prioritize prediction accuracy in strong echo regions. Thus, the common practice in radar echo extrapolation is to use a weighted Mean Absolute Error (MAE) loss:

\begin{equation}
\mathcal{L}_{\mathrm{MAE}} = \sum_{i=1}^{T^{\prime}} w_{\mathrm{MAE}} \left| \hat{\mathcal{Y}}_i - \mathcal{Y}_i \right|_1,
\end{equation}
where $\hat{\mathcal{Y}}_i$ and $\mathcal{Y}_i$ denote the predicted and ground truth radar echo images for the future time step $t+i$. $w_{\mathrm{MAE}}$ assigns region-specific weights based on echo intensity. Specifically, referencing the exponential relationship between radar reflectivity and precipitation rates, weights for high-reflectivity grid points are exponentially scaled to emphasize their impact.

Although $\mathcal{L}_{\mathrm{MAE}}$ improves on standard MAE, it remains insufficient for capturing fine echo details. We further introduce the Focal Frequency Loss (FFL)~\cite{ref40} adapted from image reconstruction, enhancing the model's capacity to match spatial structures through Fourier frequency domain analysis. The FFL is formulated as:
\begin{equation}
\begin{split}
\mathcal{L}_{\mathrm{FFL}} &= \sum_{i=1}^{T^{\prime}} \frac{1}{HW} \sum_{u,v} w_i(u,v) \\
&\qquad \cdot \left| \mathcal{F}\{\hat{\mathcal{Y}}_i\}(u,v) - \mathcal{F}\{\mathcal{Y}_i\}(u,v) \right|^2,
\end{split}
\end{equation}
where the double summation $\sum_{u,v}$ covers all discrete frequency coordinates: $u \in \{0, 1, \dots, H-1\}$ (horizontal frequencies), $v \in \{0, 1, \dots, W-1\}$. The adaptive weight matrix $w_i(u,v)$ is defined as:
\begin{equation}
\label{eq:w_uv}
w_i(u,v) = \left| \mathcal{F}\{\hat{\mathcal{Y}}_i\}(u,v) - \mathcal{F}\{\mathcal{Y}_i\}(u,v) \right|^\alpha.
\end{equation}

To focus on meteorologically significant high-frequency patterns, we compute 2D Fast Fourier Transforms (FFT) for both predicted and ground-truth composite reflectivity fields, obtaining their frequency spectra $\mathcal{F}(\hat{\mathcal{Y}})$ and $\mathcal{F}(\mathcal{Y})$. A binary mask is constructed by thresholding the magnitude spectrum $\left|\mathcal{F}(\mathcal{Y})\right|$ to retain high-frequency components. While high-frequency masking selectively enhances critical features like developing convective cloud clusters, radar echoes inherently contain complex noise components that may lead to over-penalization of high-frequency artifacts rather than genuine precipitation signals. To prioritize high-energy precipitation-related frequencies, we introduce an amplitude constraint term that weights components by their normalized target magnitude:
\begin{equation}
\begin{split}
\label{eq:w_uv_final}
w_i(u,v) &= \left| \mathcal{F}\{\hat{\mathcal{Y}}_i\}(u,v) - \mathcal{F}\{\mathcal{Y}_i\}(u,v) \right|^\alpha\\ &\cdot  \frac{|\mathcal{F}\{\mathcal{Y}_i\}(u,v)|}{\max_{u,v} |\mathcal{F}\{\mathcal{Y}_i\}(u,v)|}.
\end{split}
\end{equation}

The composite loss combines spatial and frequency domain terms:
\begin{equation}
\mathcal{L} = \mathcal{L}_{\mathrm{MAE}} + \lambda \mathcal{L}_{\mathrm{FFL}} 
\end{equation}
where $\lambda$ balances the contributions of both losses.

\section{Experiments}
In this section, we will evaluate the performance of the proposed REE-TTT model on datasets, including prediction accuracy experiments on a single dataset and cross-dataset zero-shot experiments across different regions, to demonstrate the model's strong generalization capabilities in various scenarios.
\subsection{Experiment Setting}
\subsubsection{Implementation Details}
All experiments in this chapter were implemented using the PyTorch framework on a system equipped with an AMD Ryzen 5 5600 6-core CPU and an NVIDIA GeForce RTX 4090 GPU, running Ubuntu 22.04 LTS.

We partitioned the datasets as follows:  we first isolated one complete year of precipitation data as the test set to simulate real-world deployment. The remaining samples from other years were then split into training and validation sets with an 8:2 ratio, ensuring no temporal overlap between these sets.

We set the maximum epochs to 50 with a batch size of 8. The AdamW optimizer was employed for training, starting with a learning rate of $5\times10^{-3}$ that gradually decreased to $1\times10^{-4}$. Spatial resolution was downsampled to $64\times64$ across all experiments.

\subsubsection{Baselines}
We chose ConvLSTM~\cite{ref21}, PredRNN~\cite{ref22}, MIM~\cite{ref24}, SwinLSTM~\cite{ref25} based on the recurrent neural network architecture; SIMVP~\cite{ref29}, TAU~\cite{ref30}, Earthfarseer~\cite{ref31}, which have similar spatial encoding-decoding architectures to our model; and WaST~\cite{ref35}, the latest non-recurrent neural network architecture in the field of spatio-temporal prediction, were compared with our method as baseline models. Below is a brief explanation of them: 
\begin{itemize}
\item \textit{ConvLSTM}: Integrates convolutional operations into LSTMs for spatio-temporal prediction modeling.  

\item \textit{PredRNN}: Constructs spatio-temporal LSTMs with vertical feature propagation.  

\item \textit{MIM}: Replaces LSTM forget gates with cascaded recurrent blocks to address non-stationarity.  

\item \textit{SwinLSTM}: Substitutes ConvLSTM's convolutions with Swin Transformer blocks for global spatial feature extraction.  

\item \textit{SIMVP}: A pure convolutional architecture for video prediction.  

\item \textit{TAU}: Proposes temporal attention modules for parallel evolution modeling.  

\item \textit{Earthfarseer}: Combines convolutions and Fourier Transformers for local-global interaction.

\item \textit{WaST}: Separates low-/high-frequency components via 3D wavelets and modulates them with time-frequency transformers.
\end{itemize}
\subsubsection{Evaluation Metrics}
We adopt deep learning evaluation metrics and meteorological binary classification metrics to comprehensively assess model performance. For image reconstruction quality, we employ Mean Squared Error (MSE) to quantify numerical discrepancy and the Structural Similarity Index Measure (SSIM) to evaluate perceptual similarity. For precipitation event detection, we utilize binary classification metrics based on reflectivity threshold $\tau$, including Probability of Detection (POD), False Alarm Ratio (FAR), Critical Success Index (CSI), and Equitable Threat Score (ETS), calculated as follows:

\begin{align}
&\text{POD} = \frac{\text{Hits}}{\text{Hits} + \text{Misses}}, \label{eq:pod} \\
&\text{FAR} = \frac{\text{False Alarms}}{\text{Hits} + \text{False Alarms}}, \label{eq:far} \\
&\text{CSI} = \frac{\text{Hits}}{\text{Hits} + \text{Misses} + \text{False Alarms}}, \label{eq:csi}
\end{align}
\begin{equation}
    \begin{split}
        \text{ETS} &= \frac{\text{Hits} - R}{\text{Hits} + \text{Misses} + \text{False Alarms} - R}, 
\\
R &= \frac{(\text{Hits} + \text{False Alarms})(\text{Hits} + \text{Misses})}{\text{Total Samples}}. 
    \end{split}
\end{equation}

where meteorological event categories are defined as: \text{Hits} (predicted value = true value = 1), \text{False Alarms} (predicted value = 1, true value = 0) and \text{Misses} (predicted value = 0, true value = 1). The CSI metric is reported at $\tau = {10, 25, 35}$ dBZ, corresponding to light, moderate and heavy precipitation intensities respectively. All other meteorological metrics use $\tau = 25$ dBZ. Throughout all experiments, models achieving the best validation performance on the ETS metric were selected for testing.
\begin{figure*}[!t]
\centering
\includegraphics[width=1.8\columnwidth]{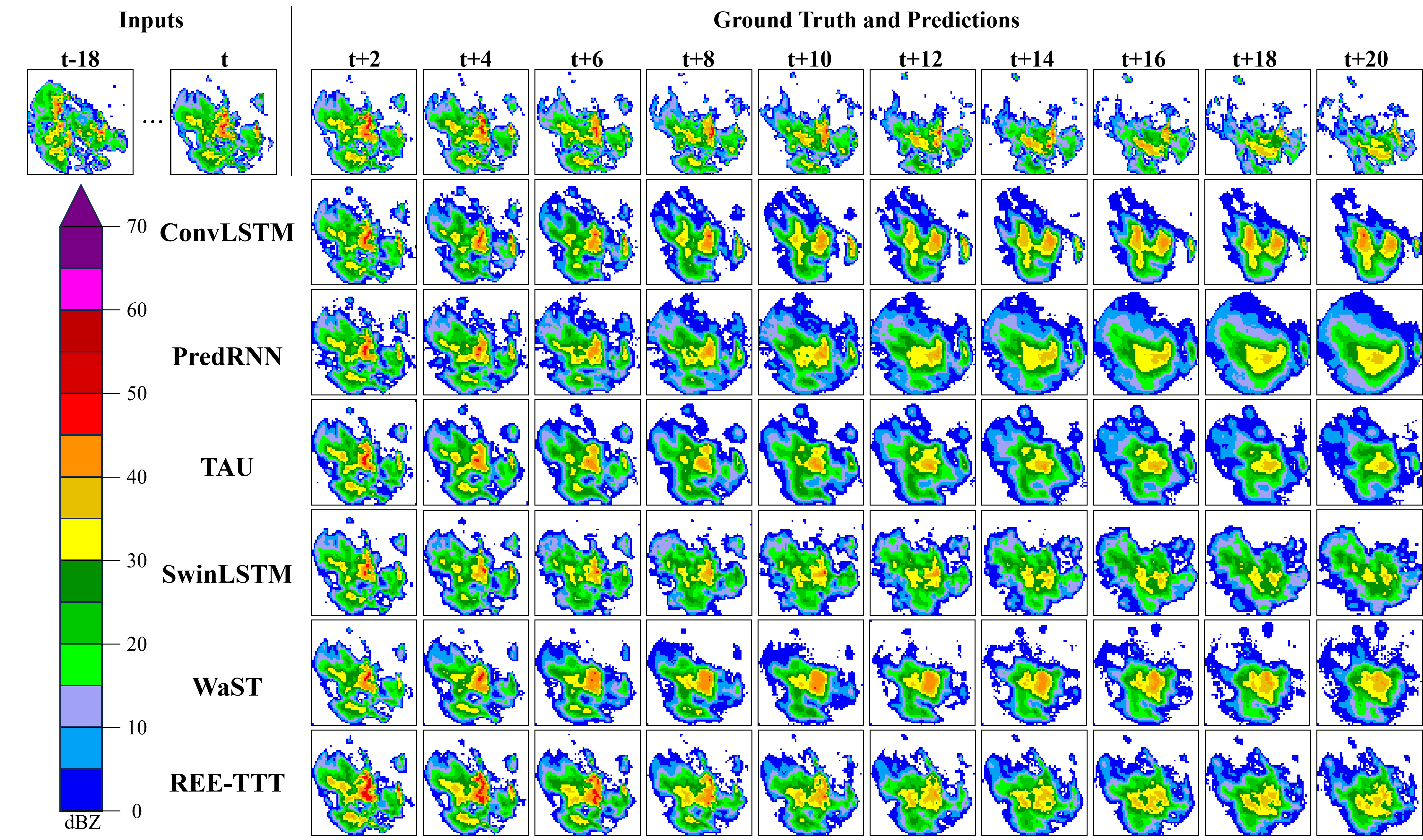}
\caption{Beijing radar echo extrapolation visualization.}
\label{fig_4}
\end{figure*}

\begin{figure*}[!t]
\centering
\subfloat[]{\includegraphics[width=1\columnwidth]{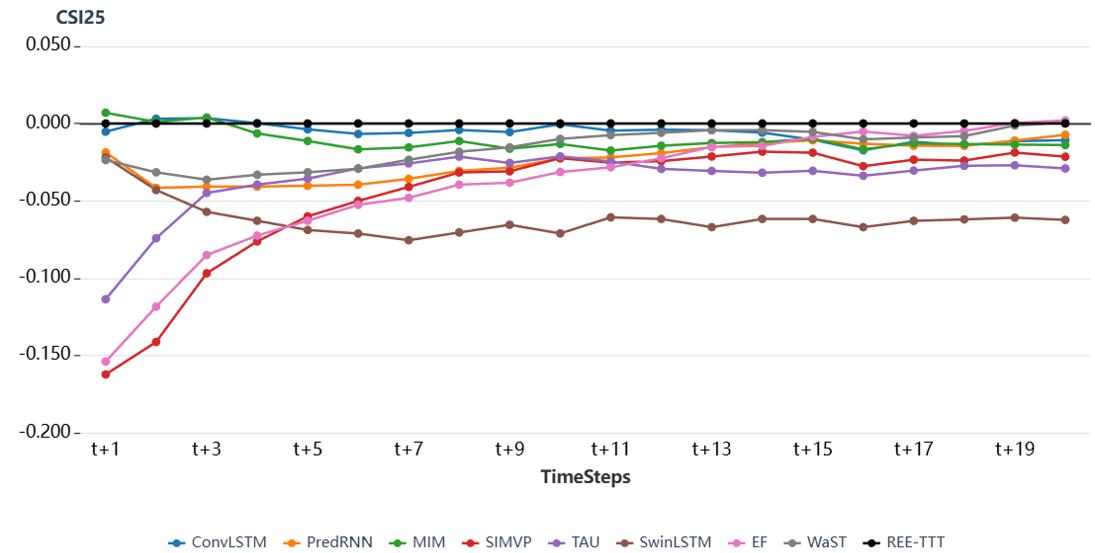}%
\label{fig5_a}}
\hfil
\subfloat[]{\includegraphics[width=1\columnwidth]{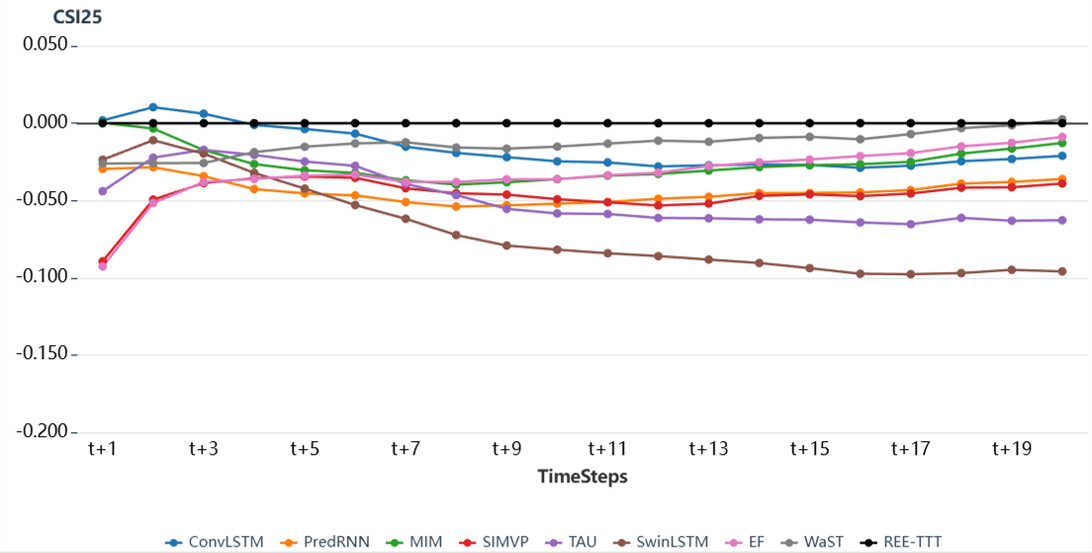}%
\label{fig5_b}}
\caption{Demonstrate the CSI$_{25}$ metric comparison across prediction timesteps, with REE-TTT serving as the baseline. Positive values indicate superior performance relative to the REE-TTT method. Subfigure (a) displays Beijing dataset results, and (b) displays Beijing-pretrained models on Hangzhou data.}
\label{fig_5}
\end{figure*}

\begin{figure*}[!t]
\centering
\subfloat[]{\includegraphics[width=1\columnwidth]{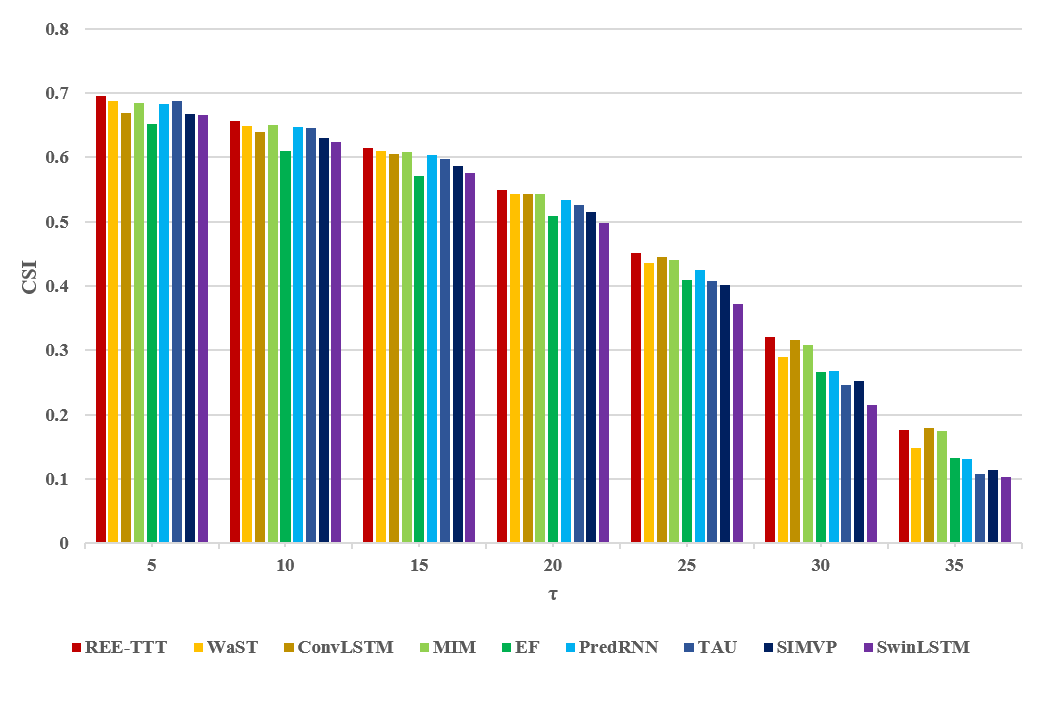}%
\label{fig6_a}}
\hfil
\subfloat[]{\includegraphics[width=1\columnwidth]{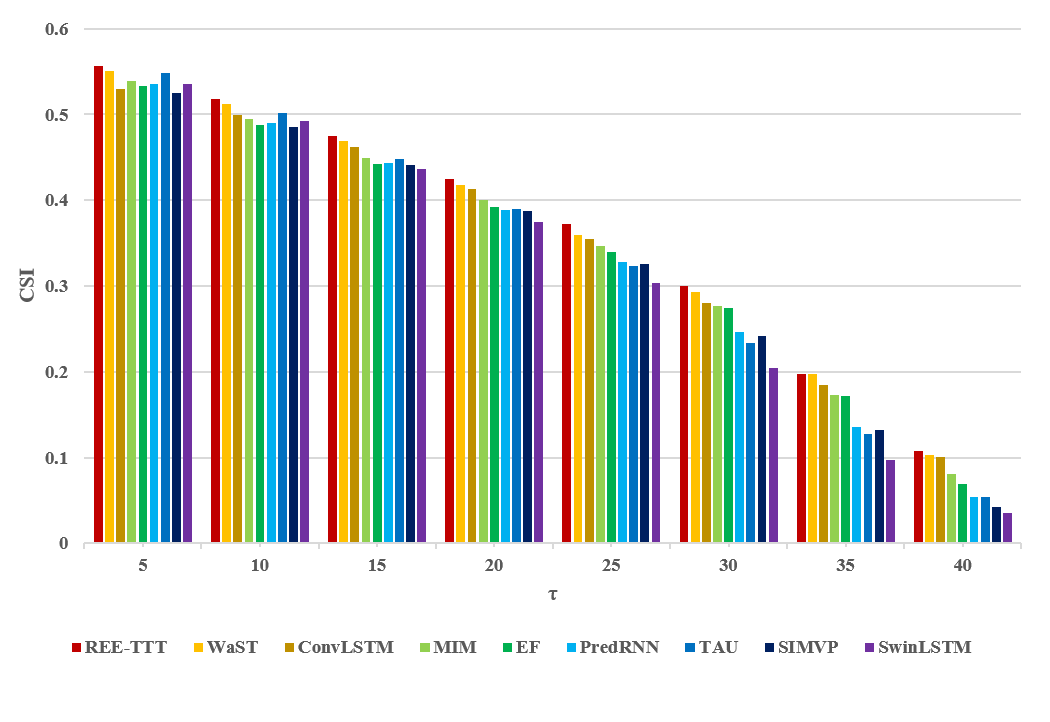}%
\label{fig6_b}}
\caption{Demonstrate CSI metric comparisons under varying binary classification thresholds $\tau$. Subfigure (a) displays Beijing dataset results, and (b) displays Beijing-pretrained models on Hangzhou data.}
\label{fig_6}
\end{figure*}

\begin{table*}[!t]
\caption{Comparative Performance Analysis\label{tab:comparative}}
\centering
\footnotesize
\begin{tabular}{cc cccccccc}
\toprule
Method & Date & \multicolumn{1}{c}{MSE$\downarrow$} & \multicolumn{1}{c}{SSIM$\uparrow$} & \multicolumn{1}{c}{FAR$\downarrow$} & \multicolumn{1}{c}{POD$\uparrow$} & \multicolumn{1}{c}{CSI$_{10}$$\uparrow$} & \multicolumn{1}{c}{CSI$_{25}$$\uparrow$} & \multicolumn{1}{c}{CSI$_{35}$$\uparrow$} & \multicolumn{1}{c}{ETS$\uparrow$} \\ 
\midrule
ConvLSTM & NIPS'15 & 21.442 & 0.749 & 0.460 & 0.662 & 0.639 & \underline{0.445} & \textbf{0.180} & \underline{0.420} \\
PredRNN & NIPS'17 & \textbf{18.243} & 0.742 & \textbf{0.422} & 0.589 & 0.648 & 0.425 & 0.132 & 0.405 \\
MIM & CVPR'19 & 20.592 & \underline{0.755} & 0.457 & 0.669 & \underline{0.650} & 0.440 & 0.175 & 0.418 \\
SimVP & CVPR'22 & 20.636 & 0.745 & 0.493 & 0.623 & 0.629 & 0.402 & 0.114 & 0.379 \\
TAU & CVPR'23 & 19.497 & 0.728 & 0.487 & 0.630 & 0.628 & 0.411 & 0.123 & 0.388 \\
SwinLSTM & ICCV'23 & \underline{19.422} & 0.740 & \underline{0.434} & 0.503 & 0.632 & 0.376 & 0.089 & 0.357 \\
EarthFarseer & AAAI'24 & 21.806 & 0.730 & 0.510 & \underline{0.679} & 0.622 & 0.419 & 0.132 & 0.384 \\
WaST & CVPR'24 & 20.972 & 0.748 & 0.484 & \textbf{0.692} & 0.643 & 0.435 & 0.149 & 0.412 \\        REE-TTT & -- & 19.457 & \textbf{0.763} & 0.453 &0.678 & \textbf{0.657} & \textbf{0.451} & \underline{0.176} & \textbf{0.427} \\
\bottomrule
\end{tabular}

\vspace{2mm}
\begin{minipage}{\dimexpr\textwidth-2\tabcolsep}
\raggedright
\footnotesize
The bast score is marked in bold, and second best score is underlined. Arrows denote optimization directions ($\uparrow$=higher better, $\downarrow$=lower better).
\end{minipage}
\end{table*}

\subsection{Comparative Experiments}  
We conduct comparative experiments using composite reflectivity radar data from the Beijing region. This dataset is derived from raw radar base data collected at Beijing Meteorological Station (Z9010) from March to September during 2010--2021, with a radar volume scan interval of 6 minutes, scanning radius of 230 km, and native resolution of $256 \times 256$ composite reflectivity images. Given Beijing's relatively low annual precipitation, we follow~\cite{ref41} to select only significant precipitation periods. Each continuous 40-frame sequence constitutes a data sample, filtered under these criteria: at least 2 frames must contain significant precipitation areas (reflectivity $\geq 25$ dBZ) covering $\geq 5\%$ of the scanning domain. The final dataset contains 639 valid sample groups, each comprising 40 consecutive frames. The first 20 frames (2 hours) are used to predict the subsequent 20 frames (2 hours), with radar echo intensities constrained between 0--70 dBZ throughout the dataset.

As shown in Table \ref{tab:comparative}, our proposed REE-TTT model demonstrates significant precision advantages in radar echo extrapolation tasks. From deep learning metrics, its SSIM outperforms all baseline models, indicating superior structural fidelity and human‑perceived quality in predictions. Although not optimal in pixel‑level error assessment (MSE), this limitation may be attributed to an inherent characteristic of the evaluation itself: given that radar echo data are dominated by vast areas of low‑intensity background, even minor deviations in these regions can disproportionately influence the MSE, potentially obscuring the model’s superior performance in capturing the critical, high‑intensity precipitation cores that are meteorologically most significant.

In meteorological event detection capabilities, REE-TTT achieves comprehensive leadership. It attains the highest POD, suggesting a tendency to predict precipitation occurrence more aggressively. While this could theoretically induce over-prediction, REE-TTT maintains average FAR levels, indicating effective balance between detection sensitivity and specificity. Consequently, REE-TTT achieves optimal scores in composite metrics: CSI and ETS. In contrast, both PredRNN and SwinLSTM achieve lower MSE through predictive smoothing strategies, which benefits global error control. However, this comes at the cost of event sensitivity -- particularly evident in SwinLSTM's poor binary classification performance. The WaST model approaches REE-TTT in POD but suffers higher FAR, indicating stronger over-prediction tendencies. We select the sample with the highest precipitation intensity in the test set for visualization, and the results are shown in Fig. \ref{fig_4}. Our model successfully predicted the weakening of the main precipitation cloud cluster moving eastward and the cloud cluster behind it moving southeastward.

We further analyze time-dependent performance through CSI$_{25}$ metrics across prediction horizons, shown in Fig. \ref{fig5_a}. RNN-based models excel in short-term forecasting (0--1h) due to their strong sequential modeling capabilities. However, performance degrades rapidly beyond 1h as fixed-size hidden states struggle to preserve long-term features, allowing no-recurrent models to surpass them. REE-TTT's adaptive hidden states enable sustained superiority, becoming the optimal model after half an hour and remaining so until the end.

Threshold-sensitive analysis Fig. \ref{fig6_a} reveals REE-TTT's robustness across precipitation intensity levels. While no-recurrent architectures suffer from information loss during spatial compression-reconstruction processes, leading to underestimation of heavy precipitation events, REE-TTT's skip connections preserving low-level features and targeted high-frequency signal recovery ensure competitive performance even at high thresholds ($\tau=35$ dBZ), matching RNN-based models.

\begin{table*}[!t]
\caption{Generalization Ability Analysis\label{tab:generalization}}
\centering
\footnotesize
\begin{tabular}{cc cccccccc}
\toprule
Method & Date & \multicolumn{1}{c}{MSE$\downarrow$} & \multicolumn{1}{c}{SSIM$\uparrow$} & \multicolumn{1}{c}{FAR$\downarrow$} & \multicolumn{1}{c}{POD$\uparrow$} & \multicolumn{1}{c}{CSI$_{10}$$\uparrow$} & \multicolumn{1}{c}{CSI$_{25}$$\uparrow$} & \multicolumn{1}{c}{CSI$_{35}$$\uparrow$} & \multicolumn{1}{c}{ETS$\uparrow$} \\ 
\midrule
ConvLSTM & NIPS'15 & 62.550 & \underline{0.561} & 0.483 & 0.494 & 0.498 & 0.354 & 0.183 & \underline{0.306} \\
PredRNN & NIPS'17 & 57.671 & 0.546 & \underline{0.480} & 0.440 & 0.489 & 0.326 & 0.134 & 0.280 \\
MIM & CVPR'19 & \underline{57.510} & \textbf{0.562} & \textbf{0.460} & 0.456 & 0.495 & 0.346 & 0.174 & 0.298 \\
SimVP & CVPR'22 & 60.309 & 0.552 & 0.528 & 0.462 & 0.485 & 0.324 & 0.130 & 0.275 \\
TAU & CVPR'23 & 57.774 & 0.532 & 0.489 & 0.446 & 0.491 & 0.327 & 0.126 & 0.283 \\
SwinLSTM & ICCV'23 & \textbf{55.021} & 0.531 & 0.492 & 0.391 & 0.492 & 0.304 & 0.098 & 0.261 \\
Earthfarseer & AAAI'24 & 68.471 & 0.530 & 0.568 & 0.580 & 0.488 & 0.339 & 0.172 & 0.282 \\
WaST & CVPR'24 & 63.872 & 0.519 & 0.574 &\textbf{0.657} & \underline{0.510} & \underline{0.358} & \underline{0.196} & 0.296 \\
REE-TTT & -- & 59.155 & 0.540 & 0.528 & \underline{0.596} & \textbf{0.516} & \textbf{0.368} & \textbf{0.200} & \textbf{0.315}  \\
\bottomrule
\end{tabular}
\end{table*}

\begin{figure*}[!t]
\centering
\includegraphics[width=1.8\columnwidth]{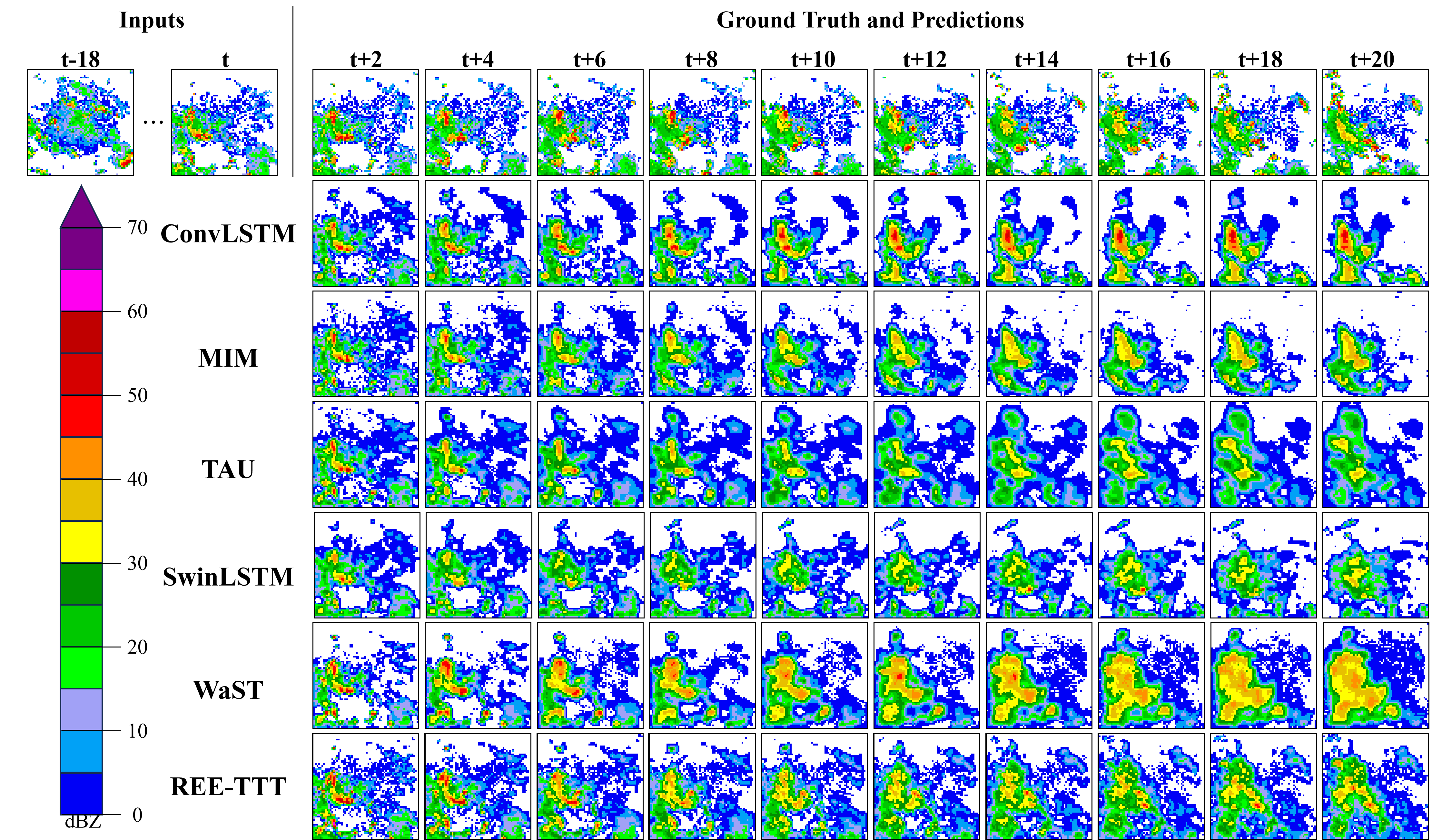}
\caption{The comparative visualization of predictions from several pretrained models for a typical severe convective event in Hangzhou shows that the REE-TTT model more accurately tracked the intense precipitation cloud clusters compared to other models, as evidenced by the marks in the final timestep.}
\label{fig_7}
\end{figure*}

\subsection{Cross-Regional Generalization Experiments}  

To validate model generalization across diverse data distributions, we test models trained on the Beijing dataset (Section 4.2) against Hangzhou's heavy precipitation scenarios. The Hangzhou dataset comprises composite reflectivity data from Hangzhou Meteorological Station (Z9571) during 2016--2018,  where heavy precipitation samples are defined by maximum echo intensities exceeding 40 dBZ. With 6‑minute temporal resolution and 0.01° spatial resolution, we extract 985 sample groups by applying a sliding window of 40 consecutive frames with a stride of 2 frames over precipitation events, thereby ensuring temporal independence between adjacent samples.Among these, 316 samples from the year 2018 are held out for zero-shot testing.

As shown in Table \ref{tab:generalization}, models demonstrate pronounced predictive biases in the cross-regional generalization experiment. Consistent with Beijing results but more pronounced here, RNN-based models (PredRNN, MIM, SwinLSTM) maintain low MSE and FAR at the cost of substantially reduced POD – indicating overly conservative predictions that systematically underdetect precipitation events. Our REE-TTT optimally balances these metrics, achieving superior CSI and ETS scores with more pronounced advantages in generalization tests. Fig. \ref{fig5_b} and Fig. \ref{fig6_b} illustrate temporal and threshold-wise performance, while Fig. \ref{fig_7} provides visual comparisons. ConvLSTM shows accurate intensity estimation but significant positional drift over time. WaST excels in intensity  prediction but over-expands strong echo regions. In contrast, REE-TTT maintains precise morphology tracking throughout prediction horizons (particularly for persistent heavy echoes), demonstrating test-time training's efficacy in cross-distribution adaptation for extreme precipitation scenarios.  

\subsection{Ablation Study and Extension Study}  
\begin{table*}[!t]
\caption{Ablation Study on Beijing and Hangzhou Datasets\label{tab:ablation}}
\centering
\footnotesize
\begin{tabular}{c *{4}{c} *{4}{c}}
\toprule
\multirow{2}{*}{Method} & \multicolumn{4}{c}{Beijing Dataset} & \multicolumn{4}{c}{Hangzhou Dataset (Zero-shot)} \\
\cmidrule(lr){2-5} \cmidrule(lr){6-9}
 & CSI$_{10}$$\uparrow$ & CSI$_{25}$$\uparrow$ & CSI$_{35}$$\uparrow$ & ETS$\uparrow$ & CSI$_{10}$$\uparrow$ & CSI$_{25}$$\uparrow$ & CSI$_{35}$$\uparrow$ & ETS$\uparrow$ \\
\midrule
No HFFL loss & 0.654 & 0.438 & 0.172 & 0.417 & 0.504 & 0.348 & 0.180 & 0.294 \\
No skip connect & 0.645 & 0.434 & 0.166 & 0.410 & 0.505  & 0.364 & 0.175 & 0.302 \\
Linear proj & 0.653 & 0.449 & 0.174 & 0.424  & 0.510 & 0.360  & 0.182 &  0.306 \\
No RRDB & \underline{0.655} & \underline{0.447} & \textbf{0.178} & \underline{0.425} & \textbf{0.516} & \underline{0.365} & \textbf{0.202} & \underline{0.311} \\
REE-TTT & \textbf{0.657} & \textbf{0.451} & \underline{0.176} & \textbf{0.427} & \textbf{0.516} & \textbf{0.368} & \underline{0.200} & \textbf{0.315} \\
\bottomrule
\end{tabular}

\end{table*}

This section investigates the contribution of individual model components through ablation experiments and validates the framework's extensibility through a three-stage learning strategy.

We progressively remove components from the complete REE-TTT model to investigate the effectiveness of each module. Specifically, we conduct four ablation configurations: reverting the attention-based projections in TTT layers to initial linear projections (denoted as \textit{Linear proj}), disabling the skip connection branches (denoted as \textit{No skip connect}), removing the frequency-domain loss (denoted as \textit{No HFFL loss}), and removing the super-resolution refinement branch (denoted as \textit{No RRDB}).

The experimental results are shown in Table \ref{tab:ablation}. The exclusive reliance on pixel-wise loss (\textit{No HFFL loss}) significantly disrupts the training convergence direction, leading to reduced model accuracy. The absence of direct low-level feature reconstruction (\textit{No skip connect}) prevents the model from effectively capturing distinctions between different echo sequences, while exclusive dependence on decoder-based reconstruction causes loss of foundational information, resulting in substantial underestimation of heavy precipitation regions. Although the linear projection variant (\textit{Linear proj}) achieves comparable accuracy to our reported model, its simplified projection scheme degrades the modeling capacity of TTT layers. The insufficient feature decoupling across different views in this configuration severely limits cross-regional generalization capability. Notably, the removal of the super-resolution branch (\textit{No RRDB}) yields quantitative metrics that are competitive with the full model. However, predictions from this configuration exhibit excessive smoothing, particularly within extreme precipitation cores. This results in blurred boundaries and dispersed intensity estimates for heavy rainfall regions. While pixel-wise statistical differences may be marginal, such loss of fine-scale structure is detrimental for operational nowcasting that demands precise localization and morphology of convective cells. The full REE-TTT configuration, integrating all components, achieves optimal performance across evaluation metrics, thereby conclusively validating both the effectiveness and rationality of our proposed methodology.

To validate the extensibility of the REE-TTT framework for adapting to new regions with distinct precipitation characteristics, we conduct an extension study employing a three-stage learning strategy. The process is structured as follows: In Stage 1 (Pre-training), the complete REE-TTT backbone—comprising the encoder, ST-TTT translator, and decoder—is trained on the source Beijing dataset. Subsequently, in Stage 2 (Fine-tuning), the pre-trained backbone parameters are frozen, and the skip connection branch (described in Section III.B, containing motion and temporal attention modules) is fine-tuned on the 2016–2017 Hangzhou dataset, alongside the inner-loop adaptive parameters within the ST-TTT blocks. A newly introduced learnable fusion weight is optimized to balance the contributions from the frozen translator and the fine-tuned skip connection path. Finally, Stage 3 (Testing) evaluates the adapted model on the same Hangzhou test set used in the zero-shot experiment.
The results are presented in Table \ref{tab:three_stage_results}. The three-stage fine-tuning strategy yields consistent improvements over the zero-shot baseline across all metrics. This demonstrates the practical extensibility of the REE-TTT framework when a limited amount of labeled data from a target domain is available for adaptation, affirming its value for cross-regional deployment.

\begin{table*}[!t]
\centering
\footnotesize
\caption{Three-Stage Fine-tuning Performance on Hangzhou Extreme Precipitation Dataset}
\label{tab:three_stage_results}
\begin{tabular}{c*{8}{c}}
\toprule
Method & MSE$\downarrow$ & SSIM$\uparrow$ & FAR$\downarrow$ & POD$\uparrow$ & CSI$_{10}$$\uparrow$ & CSI$_{25}$$\uparrow$ & CSI$_{35}$$\uparrow$ & ETS$\uparrow$ \\
\midrule
Zero-Shot & 59.155 & 0.540 & 0.528 & 0.596 & 0.516 & 0.368 & 0.200 & 0.315 \\
Three-Stage & \textbf{58.685} & \textbf{0.551} & \textbf{0.527} & \textbf{0.606} & \textbf{0.528} & \textbf{0.377} & \textbf{0.210} & \textbf{0.320} \\
\bottomrule
\end{tabular}
\end{table*}

\section{Conclusion and Limitations}
In this work, we introduce the TTT strategy into radar echo extrapolation tasks and propose the REE-TTT model based on the TTT layer. By replacing unified linear projections in standard TTT layers with differentiated attention mechanisms that model temporal evolution, our method achieves functional decoupling of feature views, enabling more effective self-supervised reconstruction tasks for TTT and significantly improving model generalization. While maintaining superior prediction accuracy, REE-TTT adaptively adjusts to test sets with distribution shifts, demonstrating clear advantages over existing methods in cross-regional extreme weather scenarios. This breakthrough allows radar echo extrapolation models pretrained on high-quality datasets to serve more precipitation scenarios, providing high-resolution forecasts for regions lacking historical precipitation data. Furthermore, the REE-TTT architecture holds potential for broader applications in spatio-temporal prediction tasks.

However, REE-TTT exhibits some limitations. First, despite targeted design improvements, the upsampling process introduces averaging effects when reconstructing spatial details of localized heavy precipitation cores, resulting in blurred predictions. This issue may become more pronounced in scenarios with higher image spatial resolution. Second, constrained by the diversity of spatio-temporal attention modules, this study primarily explores combinations of basic attention mechanisms, leaving more sophisticated designs unexplored. Future work should conduct systematic evaluations to identify optimal spatio-temporal feature interaction paradigms.

\vspace{11pt}
\begin{IEEEbiography}[{\includegraphics[width=1in,height=1.25in,clip,keepaspectratio]{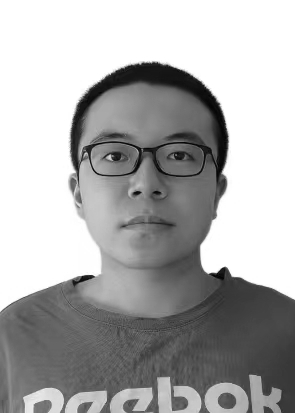}}]{Xin Di}
He is currently pursuing a master's degree in the Department of Intelligence and Automation at Beijing University of Technology. His research interests focus on spatiotemporal forecasting.
\end{IEEEbiography}

\begin{IEEEbiography}
[{\includegraphics[width=1in,height=1.25in,clip,keepaspectratio]{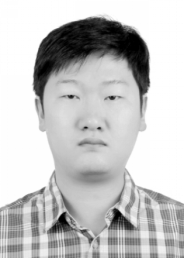}}]{Xinglin Piao}
received the Ph.D. degree from the Beijing University of Technology, Beijing, China, in 2017.

He is currently a Lecturer with the Faculty of Information Technology, Beijing University of Technology. His research interests include intelligent traffic, pattern recognition, and multimedia technology.
\end{IEEEbiography}

\begin{IEEEbiography}
[{\includegraphics[width=1in,height=1.25in,clip,keepaspectratio]{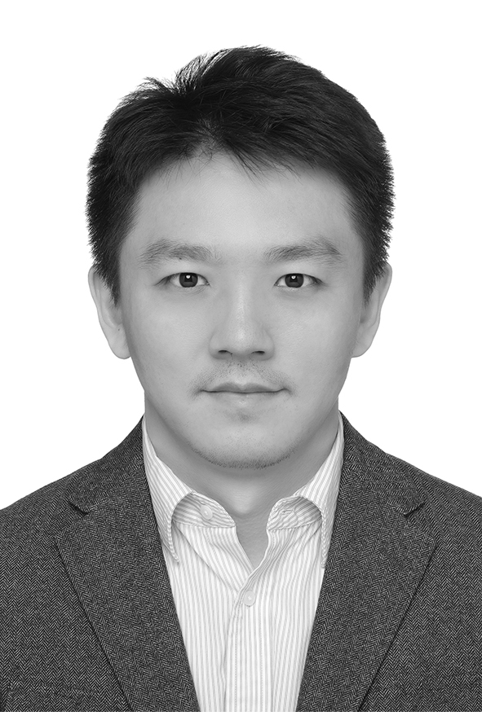}}]{Fei Wang}
received the Ph.D. degree in global environmental change from the Beijing Normal University in 2020. 

He is currently an associate research fellow at the China Meteorological Administration Weather Modification Center. His research interests include cloud physics, weather modification, and aerosol-cloud interaction.
\end{IEEEbiography}
\begin{IEEEbiography}
[{\includegraphics[width=1in,height=1.25in,clip,keepaspectratio]{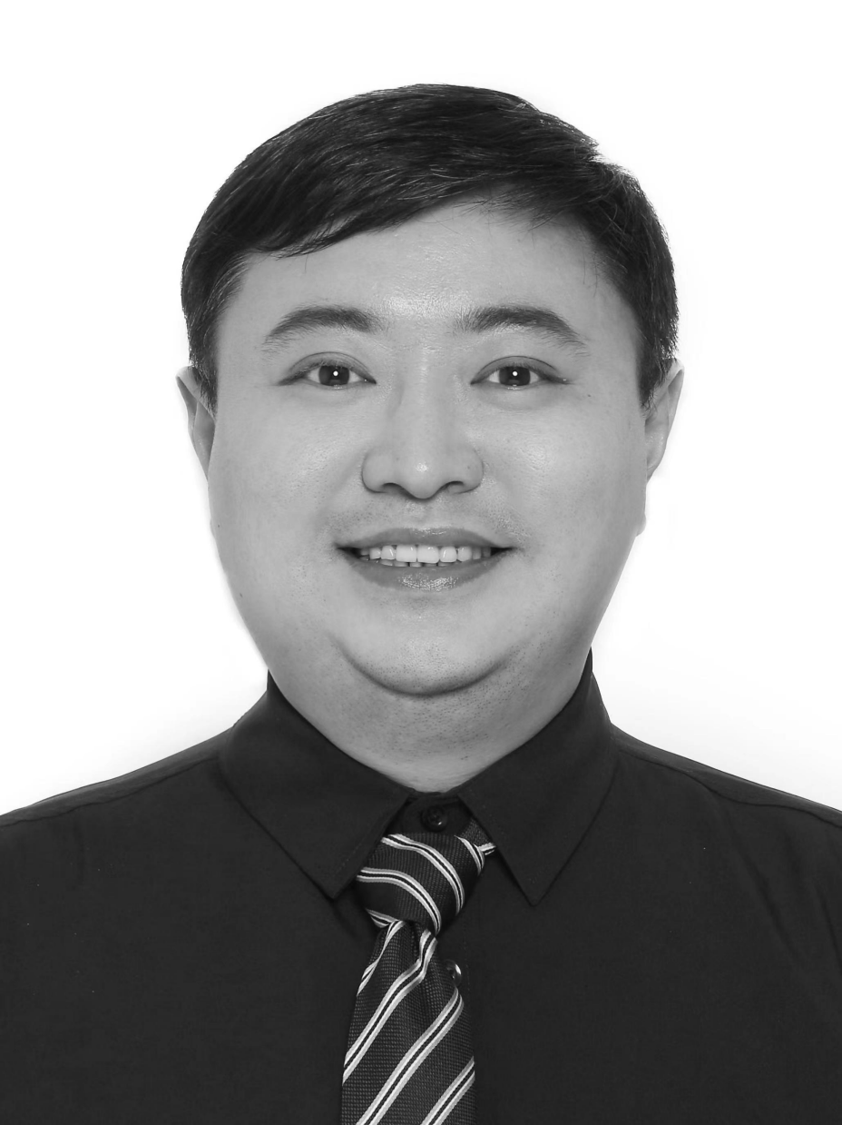}}]{Guodong Jing}
received the Ph.D. degree in computer science from the Beijing University of Technology in 2012. 

He is currently a senior principal engineer in China Meteorological Administration Weather Modification Centre. His research interests include intelligent meteorological systems, big data analysis, visualization, and computer graphics.
\end{IEEEbiography}

\begin{IEEEbiography}
[{\includegraphics[width=1in,height=1.25in,clip,keepaspectratio]{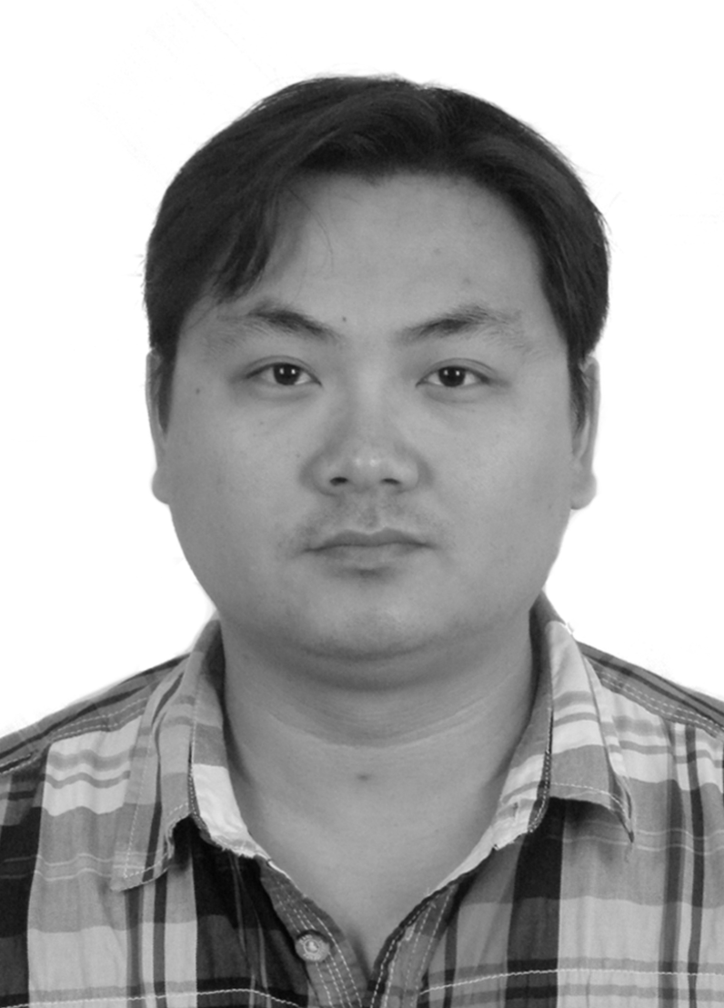}}]{Yong Zhang}
(Member, IEEE) received the Ph.D. degree in computer science from the Beijing University of Technology, Beijing, China, in 2010.

He is currently a Professor in computer science with the Beijing University of Technology. His research interests include intelligent transportation systems, big data analysis, visualization, and computer graphics.
\end{IEEEbiography}
\vfill

\end{document}